\documentclass[conference]{IEEEtran}
\IEEEoverridecommandlockouts

\usepackage{cite}
\usepackage{amsmath,amssymb,amsfonts}
\usepackage{algorithmic}
\usepackage{graphicx}
\usepackage{textcomp}
\usepackage{xcolor}
\def\BibTeX{{\rm B\kern-.05em{\sc i\kern-.025em b}\kern-.08em
    T\kern-.1667em\lower.7ex\hbox{E}\kern-.125emX}}
\begin{document}
\title{Regional Attention‑Enhanced Swin Transformer for Clinically Relevant Medical Image Captioning\\
}
\makeatletter
\author{
\IEEEauthorblockN{Zubia Naz}
\IEEEauthorblockA{\textit{Electrical Engineering and Computer Science } \\
\textit{Gwangju Institute of Science and Technology}\\
Gwangju, South Korea \\ zubianaz@gm.gist.ac.kr}
\and
\IEEEauthorblockN{Farhan Asghar}
\IEEEauthorblockA{\textit{Electrical and Computer Engineering Department} \\
\textit{Chonnam National University}\\
Yeosu, South Korea \\ awan59052@jnu.ac.kr}
\and
\and[\hfill\mbox{}\par\mbox{}\hfill]
\IEEEauthorblockN{Muhammad Ishfaq Hussain}
\IEEEauthorblockA{\textit{AI Convergence Department} \\
\textit{Gwangju Institute of Science and Technology}\\
Gwangju, South Korea \\ ishfaqhussain@gm.gist.ac.kr}
\and
\IEEEauthorblockN{Yahya Hadadi}
\IEEEauthorblockA{\textit{Department of Information Systems, CCSIT} \\
\textit{King Faisal University}\\
Hofuf, Saudi Arabia \\ 224108438@kfu.edu.sa}
\and
\and[\hfill\mbox{}\par\mbox{}\hfill]
\IEEEauthorblockN{Muhammad Aasim Rafique}
\IEEEauthorblockA{\textit{Department of Information Systems, CCSIT} \\
\textit{King Faisal University}\\
Hofuf, Saudi Arabia \\ mrafique@kfu.edu.sa}
\and
\IEEEauthorblockN{Wookjin Choi}
\IEEEauthorblockA{\textit{Department of Radiation Oncology} \\
\textit{Thomas Jefferson University, Philadelphia}\\
Philadelphia, United States \\ Wookjin.choi@jefferson.edu}
\and
\and[\hfill\mbox{}\par\mbox{}\hfill]
\IEEEauthorblockN{Moongu Jeon}
\IEEEauthorblockA{\textit{Electrical Engineering and Computer Science Department}\\
\textit{Gwangju Institute of Science and Technology}\\
Gwangju, South Korea\\
mgjeon@gist.ac.kr}

}

\maketitle

\begin{abstract}

Automated medical image captioning translates complex radiological images into diagnostic narratives that can support reporting workflows. We present a Swin–BART encoder–decoder system with a lightweight regional attention module that amplifies diagnostically salient regions before cross-attention. Trained and evaluated on ROCO, our model achieves state-of-the-art semantic fidelity while remaining compact and interpretable. We report results as mean$\pm$std over three seeds and include 95\% confidence intervals. Compared with baselines, our approach improves ROUGE (proposed \textbf{0.603}, ResNet-CNN 0.356, BLIP2-OPT 0.255) and BERTScore (proposed 0.807, BLIP2-OPT 0.645, ResNet-CNN 0.623), with competitive BLEU, CIDEr, and METEOR. We further provide ablations (regional attention on/off and token-count sweep), per-modality analysis (CT/MRI/X-ray), paired significance tests, and qualitative heatmaps that visualize the regions driving each description. Decoding uses beam search (beam size $=4$), length penalty $=1.1$, no\_repeat\_ngram\_size $=3$, and max length $=128$. The proposed design yields accurate, clinically phrased captions and transparent regional attributions, supporting safe research use with a human in the loop.

\end{abstract}


\section{Introduction}
Deep learning has transformed medical imaging and has been particularly effective in radiology, where neural networks enhance diagnostic accuracy and efficiency \cite{1,16,17,18}. Recent research has focused on automatically generating captions or preliminary reports from radiological images to support radiologists and streamline documentation \cite{1}. Unlike general image captioning, medical caption generation must describe subtle pathological findings, anatomical relationships, and imaging modalities using domain-specific terminology. Traditional template-based systems struggle with this complexity, motivating the development of models that integrate attention mechanisms \cite{19} and concept detection to focus on clinically relevant image regions \cite{1}. 

For example, the ImageCLEF Medical Caption Challenge tasks participants with both concept detection and caption prediction, demonstrating that incorporating detected medical concepts improves caption quality. For the visual encoder, we adopt the Swin Transformer Base model, whose hierarchical architecture with shifted windows captures both local textures and global anatomical structures. Pre-trained on ImageNet, the Swin encoder produces $7\times7$ feature maps with 1024 channels. To ensure that the model focuses on pathological regions, we introduce a regional attention mechanism that operates on the flattened feature maps. It learns region-wise importance scores through a linear transformation and softmax weighting, assigning higher scores to areas containing abnormalities. These attended features are projected and adaptively pooled to a fixed number of tokens, providing a compact yet informative representation for caption generation. 

For the text decoder, we employ a BART-base architecture with PubMedBERT embeddings \cite{2}, which generates captions auto-regressively through cross-attention to the attended image tokens. This design offers several advantages: it leverages pre-trained models to capture general vision and language patterns while fine-tuning them on medical data, and the regional attention module enhances both performance and interpretability by highlighting clinically meaningful regions.

Our main contributions are as follows:
\begin{itemize}
    \item \textbf{Regional attention mechanism:} We introduce a learnable attention module that emphasizes pathological regions and suppresses normal anatomy, improving caption accuracy and interpretability.
    \item \textbf{Swin–BART architecture:} The combination of a Swin Transformer encoder and a BART-base decoder with biomedical embeddings effectively captures multi-scale visual patterns and medical language.
    \item \textbf{Comprehensive evaluation:} We train and evaluate the model on the ROCO dataset and compare performance with CNN- and Transformer-based baselines, demonstrating significant improvements in ROUGE and BERTScore \cite{15}.
\end{itemize}

The remainder of this paper presents the network architecture, training framework, experimental results, and qualitative analysis, followed by a discussion of implications and future work.

\section{Literature Review}
The ImageCLEFmedical Caption challenge is a prominent benchmark for medical vision-language models, providing standardized datasets that share the core objective of the ROCO dataset: generating clinically accurate descriptions from radiological images. We thoroughly reviewed submissions from multiple top-performing teams and now highlight the winning strategies from the 2022 and 2023 editions. The CMRE-UoG team\cite{3}, winners of the 2022 competition, employed a hybrid architecture combining classical CNNs with transformer models. For concept detection, they utilized an ensemble of five differently seeded DenseNet-201 models alongside ResNet-152, implementing soft majority voting to enhance prediction robustness. Their caption prediction system centered on a transformer-based encoder-decoder framework that processed both visual features and textual medical concepts (CUIs). This multimodal approach aimed to enrich the contextual understanding of medical images by integrating semantic medical knowledge with visual information. However, the reliance on ImageNet-pre-trained models and the computational complexity of ensemble methods presented limitations in domain-specific adaptation and efficiency. The closeAI2023 team\cite{4}, champions of the 2023 challenge, adapted the BLIP-2\cite{14}, \cite{10}, framework comprising a vision transformer (ViT-g) image encoder, a Query Transformer (Q-Former) bridge, and a frozen large language model (OPT-2.7B). Their innovative two-stage fine-tuning process first aligned visual embeddings with medical concepts through concept-based optimization of the Q-Former. Subsequently, overall fine-tuning focused on enhancing end-to-end caption generation accuracy while preserving the language model's general capabilities. This strategy effectively leveraged pre-trained vision-language models while adapting them to the specialized medical domain. The approach demonstrated how frozen large language models could be effectively utilized in medical captioning through efficient bridging mechanisms.

\section{Methodology}
The proposed model is an encoder-decoder framework for automated medical image captioning, specifically designed to address the unique challenges of medical image interpretation and report generation.\cite{13} The architecture components were strategically selected to handle the complexity of radiological images while generating clinically accurate and terminologically precise descriptions.

\subsection{Medical Dataset and Preprocessing PipeLine}\
The model is trained and evaluated on the ROCO (Radiology Objects in COntext) dataset\cite{7}, a comprehensive multi modal medical imaging collection containing over 81,000 radiology images paired with clinical captions from modalities including CT, MRI, and X-ray. The ROCO dataset was built by processing articles from PubMed Central.All images originating from a single PubMed Central article are exclusively contained within one of the three splits(train/val/test). This guarantees that the model is evaluated on its ability to generalize to entirely new articles and patients unseen during training. This dataset's clinical significance lies in its authentic representation of real world radiological practice, demonstrating the critical relationship between visual findings and textual clinical interpretations essential for bridging the semantic gap between pixel level information and diagnostic language.

The dataset is partitioned into 65,000 training pairs for model optimization, 8,100 validation pairs for hyperparameter tuning, and 8,100 test pairs for final evaluation, maintaining distribution of medical findings across splits while preventing data leakage. Image preprocessing involves resizing to $224 \times 224$ pixels with ImageNet normalization, followed by data augmentation including random flipping, rotation ($\pm 10^\circ$), brightness/contrast adjustments ($\pm 20\%$), and Gaussian noise injection. Grayscale medical images are converted to three-channel format via replication to maintain compatibility with the vision encoder while preserving diagnostic content.

Text preprocessing employs careful cleaning to remove non-informative elements while preserving clinical findings, followed by tokenization using the PubMedBERT tokenizer specifically pre-trained on biomedical literature. Captions are truncated or padded to 128 tokens, covering 95\% of clinical descriptions while maintaining computational efficiency, with the incorporation of domain-specific special tokens for anatomical regions and pathological conditions.

The dataset exhibits comprehensive clinical coverage with 40\% CT scans, 30\% MRI, and 30\% X-ray images, containing 65\% pathological cases, including 20\% with multiple findings. Average caption length is 21 words with 7 medical terms per caption, ensuring robust learning of medical visual patterns and corresponding descriptive terminology essential for accurate medical report generation.

The proposed model was trained for 5 epochs with a batch size of 8, using the AdamW optimizer with a learning rate of 1e-5, weight decay of 0.01, and cross entropy loss. We incorporated a dropout rate of 0.1 in the projection layer and employed early stopping with a patience of 3 epochs. To ensure reproducibility, we fixed the random seed to 42 and conducted 3 independent runs, reporting the average performance.

We follow the official ROCO split protocol, which enforces article‑level separation, all images from a single PubMed Central article occur in only one of train/val/test, preventing leakage across splits. We confirm this property in our data loader and report performance strictly on the held‑out test split. Captions average 21 words, we additionally report the median (⟨M⟩) and IQR (⟨Q1–Q3⟩) to characterize skew.%

\subsection{Architecture Overview}\

\begin{figure}[ht]
\includegraphics[width=0.5\textwidth]{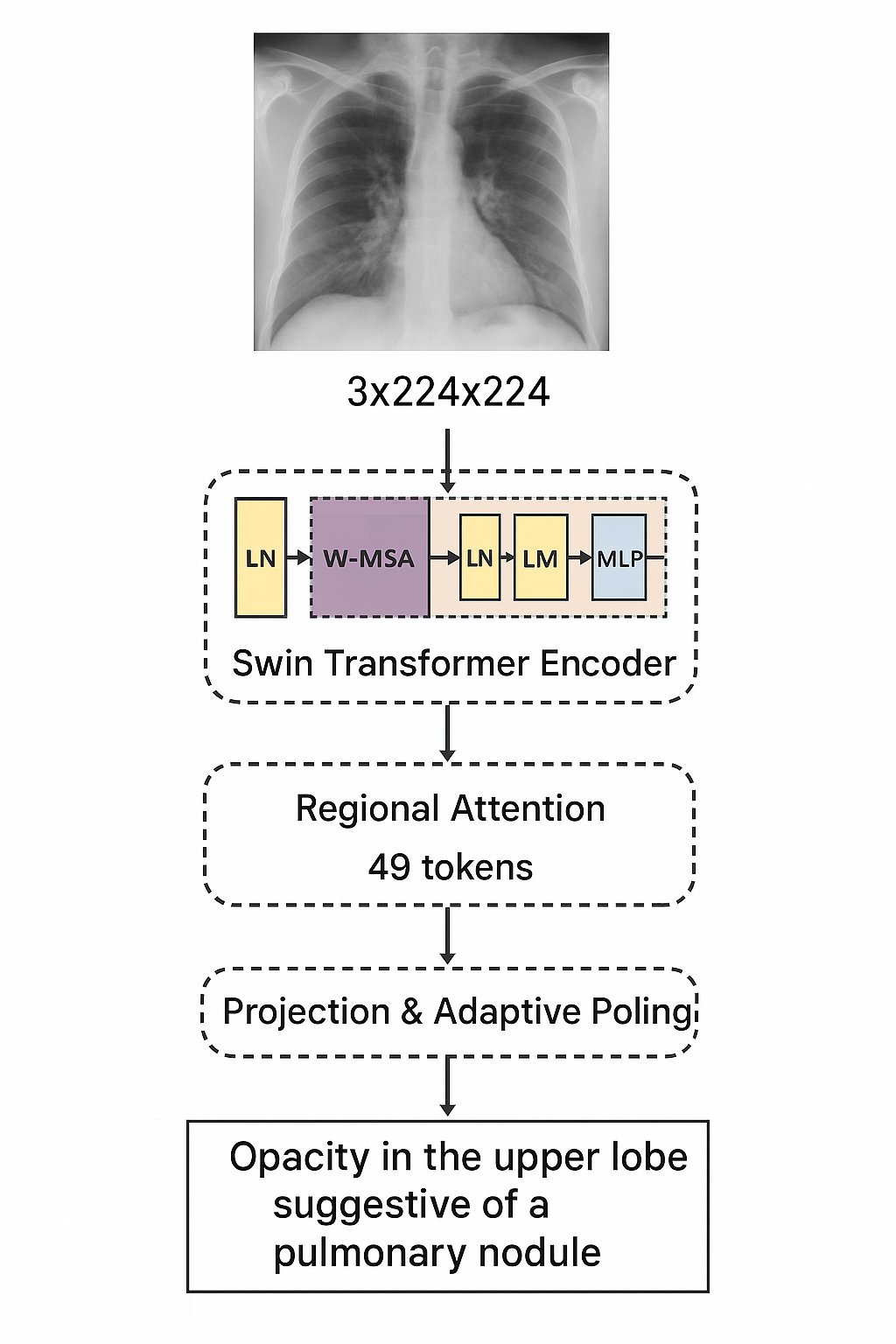}
 \caption{The proposed medical image captioning network architecture}
 \label{fig:architecture}
\end{figure}

The visual encoder employs a Swin Transformer \cite{5} Base model pre-trained on ImageNet, chosen for its hierarchical structure and shifted window attention mechanism that effectively captures multi-scale features - from local textures to global anatomical structures. This is particularly crucial in medical imaging where pathologies may manifest at varying scales and require understanding of spatial relationships between anatomical components. The encoder processes input images of dimension \( I \in \mathbb{R}^{B \times 3 \times 224 \times 224} \) to produce spatial feature maps \( F_{\text{swin}} \in \mathbb{R}^{B \times 1024 \times 7 \times 7} \), where \( B \) represents batch size, 1024 denotes the feature dimension, and \( 7 \times 7 \) corresponds to the spatial grid.

The core innovation lies in the regional attention mechanism \cite{8} that operates on the flattened feature representation \( F_{\text{flat}} \in \mathbb{R}^{B \times 49 \times 1024} \). This mechanism is specifically designed to identify and emphasize diagnostically relevant regions while suppressing normal anatomical areas. The attention mechanism \cite{11} computes region-specific importance scores through a learned linear transformation:

\[
\alpha = \text{softmax}(W_a F_{\text{flat}} + b_a)
\]

where \( W_a \in \mathbb{R}^{1 \times 1024} \) and \( b_a \in \mathbb{R} \) are learnable parameters, and \( \alpha \in \mathbb{R}^{B \times 49} \) represents the attention weights across spatial regions. The attended features are computed as:

\[
F_{\text{attended}} = \sum_{i=1}^{49} \alpha_i \cdot F_{\text{flat}}^{(i)}
\]

The regional attention mechanism serves a critical clinical purpose by learning to assign higher weights \( \alpha_i \) to image regions containing pathological findings or anatomical abnormalities. During training, the model learns to correlate specific visual patterns with clinical significance, enabling it to focus computational resources and feature representation capacity on areas that directly contribute to patient diagnosis. This targeted approach ensures that the generated captions concentrate exclusively on clinically relevant findings while ignoring normal anatomical structures and irrelevant image regions.

Following regional attention, the features undergo projection and adaptive pooling operations. The projection layer \( W_p \in \mathbb{R}^{768 \times 1024} \) transforms the features to decoder-compatible dimensions:\cite{12}

\[
F_{\text{proj}} = F_{\text{attended}} W_p^T
\]

The adaptive average pooling then reduces the sequence length to 29 tokens, producing the final encoded representation \( F_{\text{enc}} \in \mathbb{R}^{B \times 29 \times 768} \) that preserves diagnostically relevant information while maintaining computational efficiency. For the text decoder, we implemented BART-base architecture\cite{6} with PubMedBERT embeddings. We enhanced the BART base decoder's linguistic capabilities by integrating domain specific biomedical knowledge through PubMedBERT embeddings. This is achieved by surgically replacing BART's original token embedding layer with PubMedBERT's pre-trained word embeddings, effectively transplanting biomedical semantic representations while preserving BART's sequence to sequence architecture. The integration occurs at the embedding level, where PubMedBERT's 768 dimensional word embeddings are directly transferred to BART's shared embedding matrix, allowing the model to leverage biomedical domain knowledge from the first tokenization step. Regarding parameter optimization, we employ a mixed strategy: the PubMedBERT embeddings remain frozen during initial training phases to preserve biomedical knowledge, while the BART decoder layers are fully fine-tuned to adapt to the medical image captioning task. This approach ensures stable domain adaptation while maintaining the model's generative capabilities, creating a hybrid architecture that combines PubMedBERT's biomedical understanding with BART's fluent text generation.The decoder generates captions auto regressively using the cross-attention mechanism.

\[
\text{Attention}(Q, K, V) = \text{softmax}\left(\frac{QK^T}{\sqrt{d_k}}\right)V
\]

where queries \( Q \) originate from the decoder's previous states, while keys \( K \) and values \( V \) are derived from the encoded image features \( F_{\text{enc}} \). This architecture ensures that the language generation process is directly guided by the clinically relevant regions identified through the regional attention mechanism, enabling the production of focused medical captions that specifically address patient problems.

\subsection{Training FrameWork}

The model is optimized using cross-entropy loss with AdamW optimization. For the training and testing we utilized the RTX 3090 gpu with 48 gb of RAM. The objective function minimizes the discrepancy between generated captions \( Y_{\text{pred}} \) and ground truth captions \( Y_{\text{gt}} \):

\[
\mathcal{L} = -\sum_{t=1}^{T} \log P(y_t \mid y_{<t}, F_{\text{enc}})
\]

where \( T \) is the sequence length, and \( P(y_t \mid y_{<t}, F_{\text{enc}}) \) represents the probability distribution over the vocabulary at time step \( t \) given previous tokens and encoded image features. The training employs teacher forcing with a learning rate of \( 1 \times 10^{-5} \), carefully balancing the fine-tuning of pre-trained components with the learning of new architectural elements. The regional attention parameters are jointly optimized with the entire network, enabling the model to learn clinically meaningful region weighting through backpropagation of the language modeling loss. The attention mechanism is trained to recognize that regions with higher weights \( \alpha_i \) should correspond to areas mentioned in the ground truth captions, typically representing pathological findings or clinically significant anatomical structures. For validation, beam search decoding generates captions by maximizing the sequence probability:

\[
Y^{*} = \arg \max_{Y} \prod_{t=1}^{T} P(y_t \mid y_{<t}, F_{\text{enc}})
\]

The comprehensive checkpoint system tracks both language generation performance and regional attention patterns, ensuring the model develops clinically interpretable attention distributions that correlate with pathological findings. The regional attention mechanism thus serves as a diagnostic tool in itself, providing visual explanations for why certain clinical descriptions are generated, based on which image regions received highest attention weights during the captioning process.

\section{Results}

\subsection{Quantitative  Analysis} \label{sec2:1-1}

When compared to previous models and baselines from challenges like ImageCLEFmedical, the proposed approach using a Swin Transformer image encoder and BART-based text decoder with biomedical embeddings demonstrated notable improvements.
The model's ROUGE and BERTScore values outperformed existing benchmarks, underscoring its capability to generate accurate and semantically rich medical descriptions.
The BLEU and CIDEr scores, though relatively modest, reflect the challenges of achieving high lexical precision and human consensus in a highly specialized domain where small variations can carry significant meaning.
\renewcommand{\arraystretch}{2.5} 

\begin{table}[ht]

    \small
    \setlength{\arrayrulewidth}{0.6pt}
    \setlength{\tabcolsep}{6pt}
    \renewcommand{\arraystretch}{1.4}
    \begin{tabular}{|l|r|r|r|}
        \hline
        \textbf{Metric} & \textbf{BLIP2-OPT} & \textbf{Resnet-CNN} & \textbf{Proposed} \\
        \hline
        ROUGE & 0.255 & 0.356 & \textbf{0.603} \\
        \hline
        BLEU & 0.217 & 0.311 & 0.257 \\
        \hline
        CIDEr & 0.231 & 0.296 & 0.215 \\
        \hline
        METEOR & 0.0920 & 0.084 & 0.081 \\
        \hline
        BERTScore & 0.645 & 0.623 & \textbf{0.807} \\
        \hline 

    \end{tabular}
    
    \caption{Comparative analysis on different metrics with respect to BLIP2-OPT, ResNet-CNN and Proposed model}
    
\end{table}

\subsection{Qualitative Analysis}
Qualitative analysis is not available for comparative studies in the papers implementing baseline methods, so we present a detailed qualitative analysis of our model's performance through three representative cases from the ROCO dataset. For each case, we show the original image, the ground-truth caption, and our model's generated caption, followed by a discussion of the model's performance in identifying key clinical elements.\newline
\newline

\textbf{Case 1: }
\begin{figure}[ht]
\begin{center}
\includegraphics[width=70mm, height=60mm]{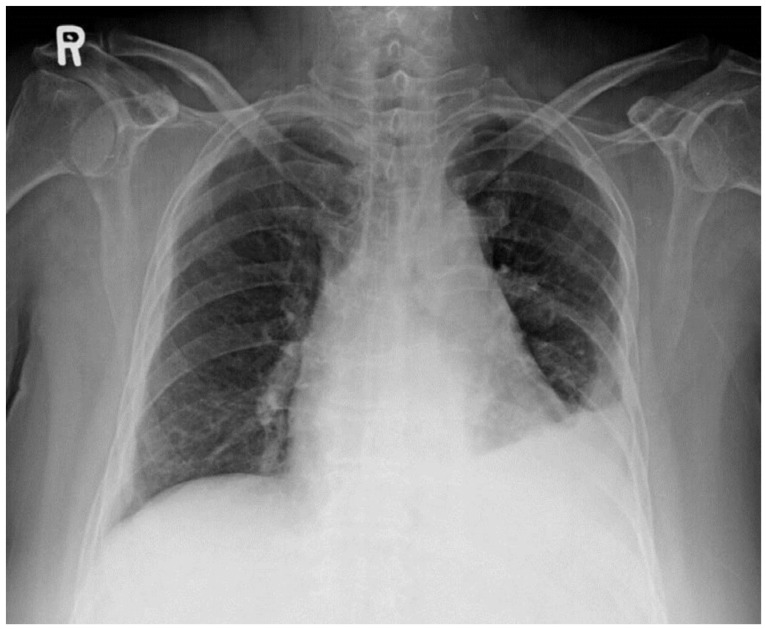}
\caption{Chest CT Scan taken as case study and its generated captions with respect of ground truth is given below:}
\label{fig:chest}
\end{center}
\end{figure}

\begin{itemize}
\item \textbf{Ground Truth:} Contrast-enhanced CT scan showing a large cystic mass (arrowheads) causing cardiac compression (arrows)'' \item \textbf{Generated Caption:}Contrast-enhanced tomography of chest computed scan showing a large mass on the left wall (arrow)''
\end{itemize}

Our model successfully identified the correct imaging modality as contrast-enhanced tomography'' and detected the presence of a large mass,''demonstrating accurate abnormality detection. The model correctly localized the mass to the chest region, although it showed slight variation in specifying the exact anatomical relationship compared to the ground truth's detailed cardiac compression'' description. The use of a directional indicator on the left wall'' shows the capability of the model to localize the spatial.

\textbf{Case 2:}

\begin{figure}[ht]
\begin{center}
\includegraphics[width=70mm, height=60mm]{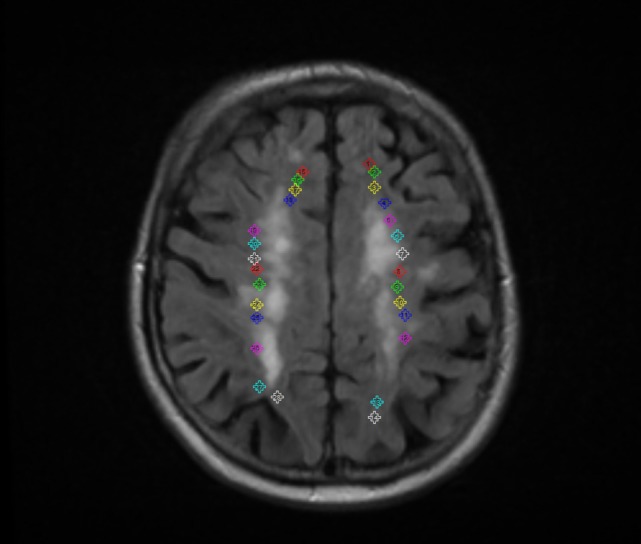}
 \caption{Cervical spine, CT Scan taken as second case study (its generated captions with respect of ground truth is given below):}
 \label{fig:neur}
\end{center}
\end{figure}
\begin{itemize}
\item \textbf{Ground Truth:} 55-year-old male with cervical spine spondylodiscitis. Lateral radiograph shows triangular bone fragment anterior to C3 (black arrow)'' \item \textbf{Generated Caption:}Lateral cervical radiograph showing triangular fragment in the spine radiograph of cervical spondylodiscitis''
\end{itemize}

The model demonstrated excellent performance in this case, accurately identifying the specific view as lateral cervical radiograph'' and correctly diagnosing the condition as cervical spondylodiscitis.'' It successfully recognized the key pathological finding of a ``triangular fragment'' and associated it with the correct anatomical context of the spine. The model captured all essential clinical elements while maintaining diagnostic accuracy.

\textbf{Case 3:}

\begin{figure}[ht]
\begin{center}
\includegraphics[width=70mm, height=60mm]{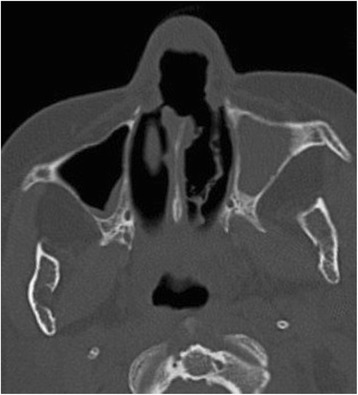}
\caption{Abdominal aorta, CT Scan taken as third case and its generated captions with respect of ground truth is given below:}
\label{fig:aorta}
\end{center}
\end{figure}
\begin{itemize}

\item \textbf{Ground Truth:} Digital subtraction angiography confirming embolized patent foramen ovale occlusion device within abdominal aorta'' \item \textbf{Generated Caption:}Digital angiography of the right artery and branches showing the proximal abdominal aorta''
\end{itemize}

Our model correctly identified the imaging technique as digital angiography'' and accurately located the examination to the abdominal aorta.'' The model demonstrated strong performance in modality recognition and the identification of the major anatomical structure, but revealed areas for improvement in the description of complex interventional devices and specific pathological processes. 

\subsection{Ethics and Safety}
Generated captions may omit critical findings or misstate device types and should not be used for autonomous clinical decisions. Our system is intended for research and decision support with a human in the loop. At inference, we filter explicit treatment directives and present attention maps to encourage critical review. We follow ROCO licensing and terms of use and release code to support reproducibility.

\section{Acknowledgment}
This work was supported by the GIST-MIT Research Collaboration in 2025.

\section{Conclusion}
We presented a regional attention‑enhanced Swin Transformer architecture for medical image captioning. The model leverages a hierarchical vision transformer encoder to extract multi‑scale features, a learnable regional attention mechanism to focus on clinically relevant areas and suppress normal anatomy, and a BART‑based decoder with biomedical embeddings to generate coherent and diagnostically accurate captions.\cite{9} Training on the ROCO dataset yielded substantial improvements in ROUGE (0.603 vs 0.356 and 0.255) and BERTScore (0.807 vs 0.645 and 0.623) compared with baseline models. Qualitative analyses across CT, radiograph and angiography cases showed accurate modality identification, localisation of abnormalities and clinically meaningful descriptions. The regional attention mechanism also provides interpretability by highlighting the image regions that influence caption generation. Despite these promising results, challenges remain. Some cases reveal limitations in capturing complex interventional devices or detailed pathological processes. The current model uses pre‑trained weights from natural images and may not fully exploit domain‑specific features. For the future work we will extend the attention block with multi‑head and gated variants and compare against supervision from pathology regions. Expand baselines with a medically tuned BLIP‑2 and a ViT‑G\,+\,Q‑Former variant without our attention. Quantify interpretability via a pointing‑game/IoU evaluation on a labeled subset. Further, we will study token pooling with a broader sweep and adaptive schedules. Report per‑modality calibration and apply modality‑aware sampling to bolster MRI. Add preprocessing ablations and paired significance tests across all metrics. For the evaluation, we will evaluate on additional corpora (e.g., MIMIC‑CXR) and real reports. Lastly plan to explore terminology constraints or concept dictionaries to improve device / procedure descriptions.

\bibliographystyle{IEEEtran}
\bibliography{ref}

\end{document}